\documentclass[runningheads]{llncs}
\usepackage{graphicx}
\usepackage{tabu}                     
\usepackage{multirow}                 
\usepackage{multicol}                 
\usepackage{multirow}                
\usepackage{float}                    
\usepackage{makecell}                 
\usepackage{booktabs}                 
\usepackage{bm}
\usepackage{algorithm}
\usepackage{svg}
\usepackage{algorithmic}
\usepackage{amsmath}
\usepackage{amssymb}
\usepackage{xcolor}
\usepackage[T1]{fontenc}
\usepackage{hyperref}
\usepackage{xcolor}
\hypersetup{colorlinks=true, linkcolor=blue, citecolor=blue, urlcolor=blue}
\usepackage{marvosym}
\usepackage{latexsym}

\let\svthefootnote\thefootnote
\newcommand\freefootnote[1]{%
  \let\thefootnote\relax%
  \footnotetext{#1}%
  \let\thefootnote\svthefootnote%
}

\usepackage{perpage} 
\MakePerPage{footnote} 
\usepackage{graphicx,verbatim}
\begin{document}
\title{
Hierarchical Brain Structure Modeling for Predicting Genotype of Glioma}


%


\author{
Haotian Tang\textsuperscript{1,*} \and
Jianwei Chen\textsuperscript{3,*} \and
Xinrui Tang\textsuperscript{2} \and
Yunjia Wu\textsuperscript{1} \and
Zhengyang Miao\textsuperscript{1} \and
Chao Li\textsuperscript{2,3,4}\textsuperscript{\,\Letter}
}
\authorrunning{H. Tang et al.}
\institute{
\textsuperscript{1}College of Medicine and Biological Information Engineering, Northeastern University, Liaoning, China \\
\textsuperscript{2}School of Science and Engineering, University of Dundee, Dundee, UK \\
\textsuperscript{3}School of Medicine, University of Dundee, Dundee, UK \\
\textsuperscript{4}Department of Applied Maths and Theoretical Physics, University of Cambridge, Cambridge, UK \\
\textsuperscript{\,\Letter}Corresponding author: \email{cl647@cam.ac.uk}
}


\maketitle
\begin{abstract}
Isocitrate DeHydrogenase (IDH) mutation status is a crucial biomarker for glioma prognosis. However, current prediction  are limited by the low availability and noise of functional MRI. Structural and morphological connectomes offer a non-invasive alternative, yet existing approaches often ignore the brain’s hierarchical organisation and multiscale interactions. To address this, we propose Hi-SMGNN, a hierarchical framework that integrates structural and morphological connectomes from regional to modular levels. It features a multimodal interaction module with a Siamese network and cross-modal attention, a multiscale feature fusion mechanism for reducing redundancy, and a personalised modular partitioning strategy to enhance individual specificity and interpretability. Experiments on the UCSF-PDGM dataset demonstrate that Hi-SMGNN outperforms baseline and state-of-the-art models, showing improved robustness and effectiveness in IDH mutation prediction.


\freefootnote{$*$ Equal contribution.}
\freefootnote{Accepted at MICCAI 2025 Workshop on Computational Mathematics Modeling in Cancer Analysis}

\end{abstract}

\section{Introduction}
Glioma is one of the most common malignant brain tumors in adults, with varying prognosis~\cite{yan2024spatiotemporal}. The Isocitrate DeHydrogenase(IDH) mutation status is one of the most important biomarkers in the diagnosis and prognosis of glioma. Typically, IDH wild-type gliomas have a more favorable prognosis compared to mutant types\cite{wei2021predicting}. Current methods for determining IDH mutation status rely primarily on immunohistochemistry and targeted gene sequencing of surgical samples. However, these are invasive techniques that pose risks to the patients and are not feasible for patients who are ineligible for tumor resection or biopsy. Therefore, there is a growing interest in developing non-invasive imaging-based approaches to predict IDH mutation status and assist in glioma diagnosis.

Brain connectome, graphical representations of the brain, characterizes brain organization based on MRI~\cite{wei2021quantifying,wei2023structural}.
Specifically, the structural connectome constructed from diffusion MRI (dMRI) reflects white matter integrity between  regions~\cite{li2021brainnetgan}, while the functional connectome based on functional MRI (fMRI) reveals the co-activation of brain regions~\cite{hu2024d}. Integrating multi-modal connectome could characterize the brain more comprehensively~\cite{zhang2022predicting,chen2023group}. For instance, Ye et al.~\cite{ye2023rh} proposed a regional multimodal fusion strategy based on regional network representations. Jing et al.~\cite{xia2024img} proposed to guide the learning process with regional multimodal interactions. Despite the promising performance, combining both connectomes remains a suboptimal approach for IDH mutation status prediction, primarily due to the inherently low signal-to-noise ratio of fMRI and its limited availability in clinical glioma datasets\cite{hu2024graph}. Thus, a more precise representation is needed.

The regional radiomics similarity network (R2SN), constructed from structural MRI (sMRI) by computing inter-regional similarities of radiomic features extracted from brain regions, has been proposed as an advanced morphological connectomic framework~\cite{zhao2022regional,hu2024graph}. Since the glioma induces more pronounced and stable morphological alterations, integrating the structural and morphological connectome can effectively model the co-changed physical information of white matter fiber tracts and gray matter morphological features, thereby providing a more accurate structural and pathological modeling of the patient's brain. However, simply summarizing or concatenating both representations may not effectively reflect the complex organization of the human brain.

Neuroscience advances have revealed that the brain is hierarchically organized, while interactions between structure and morphological connectome can be observed across different levels of this hierarchy, at regional, modular, and global scales. Therefore, modelling the hierarchical interactions of the connectome could comprehensively capture the brain topology for multimodal connectome fusion. However, challenges persist in developing a hierarchical model for the brain connectome. \textbf{First}, it remains challenging to capture the precise interactions of multiscale brain connectome due to the limited receptive field, the large number of connections and noisy connectivities. \textbf{Second}, multiscale interactions of multimodal connectomes varies across individuals, requiring an adaptive approach to effectively model the interactions while reflecting individuality among the population.
\textbf{Finally}, interactions at different scales exhibit intrinsic heterogeneity. Fine-grained interactions inherently contain more redundant features. This necessitates a hierarchical fusion mechanism capable of discriminative feature refinement and cross-scale dependency integration.

To address these challenges, we propose a hierarchical framework integrating structural and morphological connectome(Hi-SMGNN). Specifically, Hi-SMGNN constructs hierarchical interaction representations from regional to modular levels and integrates the multiscale representations to obtain a global interaction representation. To effectively characterize interactions of multimodal connectomes, we propose a \textbf{multimodal interaction module} based on a Siamese network and cross-modal attention mechanism, which captures the interactions from fine-grained features by filtering noise. Additionally, given the individual variability of the human connectome, we propose an \textbf{personalized modular partitioning method} based on the topological modularity, enhancing the neuroscience interpretability of the model. Further, we develop a \textbf{multiscale features fusion mechanism} to actively suppress redundant regional features via modular features based on soft thresholding and fuses different scale interaction information effectively. Our contributions include:
\begin{itemize}
    \item A hierarchical framework integrating the structural connectome and morphological connectome to model the brain more comprehensively, enhancing IDH status prediction.
    \item A multimodal interaction module based on a Siamese network and a cross-modal attention mechanism to efficiently capture the fine-grained interaction representations.
    \item A personalized modular partitioning method based on the topological modularity to account for individual variability, enhancing the neuroscience interpretability of the model.
\end{itemize}

\begin{figure}[t]
\includegraphics[width=\textwidth]{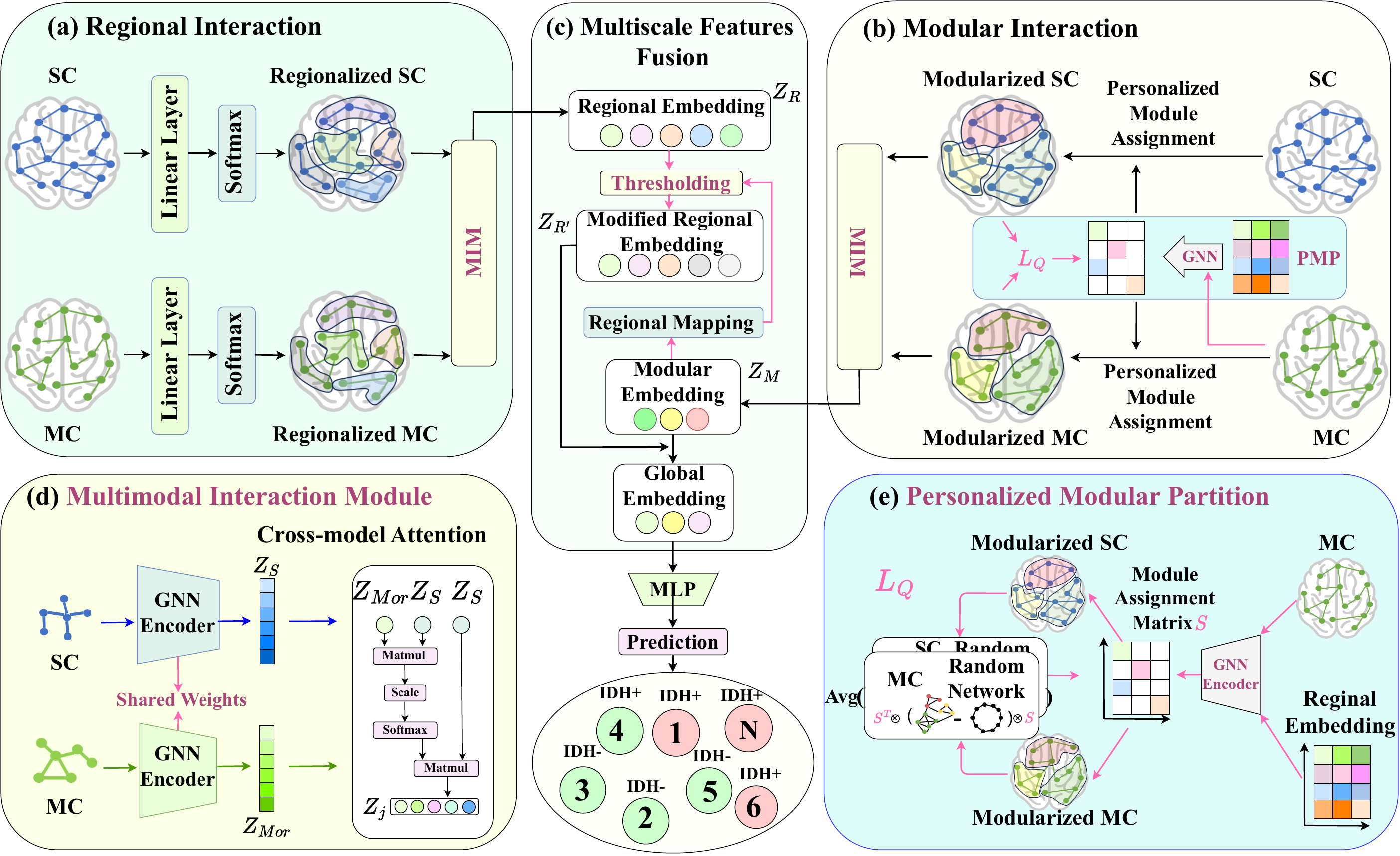}
    \caption{The structural connectivity(SC) and morphological connectivity(MC) generate (a) regionalized and  (b) modularized interaction representations through Personalized Module Partition. (c) These representations are combined to generate global interaction representations through multiscale features fusion. (d) The multimodal interaction is modelled by a Siamese network and cross-modal attention. (e) The Personalized Module Partition is achieved with a soft assignment matrix, optimized by the modularity score.} 
    \label{fig}
\end{figure}

\section{Methodology}
The proposed Hi-SMGNN (Fig.~\ref{fig}) takes the input structural connectivity(SC) and morphological connectivity(MC) to obtain corresponding regionalized SC and MC, fed into the proposed \textbf{Multimodal Interaction Module (MIM)} (Sect.~\ref{Sec.1}) to generate regionalized interaction representations. The regionalized SC and MC are further partitioned by an Personalized Module Partition to obtain modularized SC and MC, fed into MIM to obtain modularized interaction representations. Global interaction representations are yielded by fusing regionalized and modularized interaction representations through a \textbf{Multiscale Features Fusion(MFF)} mechanism (Sect.~\ref{Sec. 2}). The global interaction representations are further modelled by a Multilayer Perceptron (MLP) for downstream predictions. The proposed \textbf{Personalized Modular Partition (PMP) module} (Sec.~\ref{Sec. 3})  is achieved with a soft assignment matrix, optimized by the modularity score calculated as a connectome topological feature.

\subsection{Preliminaries }Given a multimodal connectome dataset $
\mathcal{D}=\left\{ \mathcal{P}_1,...,\ \mathcal{P}_P \right\}$  that contains $P$ subjects, each with a paired SC  ($G_S$) derived from dMRI and MC ($G_{Mor}$) derived from sMRI, and a label $y$ determined by the IDH status. Each type of graph $G=\left( V,X,A \right)$ consists of a node set $V$, an adjacency matrix $A\in \mathbb{R}^{N\times N}$ and node features $X\in \mathbb{R}^{N\times d}$, where $N=|V|$ represents the number of nodes (brain regions) and $d$ represents the dimension of node features. In this study, the structural and radiomics node features $\left( X_S,\ X_{Mor} \right)$ are initialized by the corresponding rows of the adjacency matrix for that type. The aim of the multimodal connectome fusion is to find a mapping function $f:\left( G_S,\ G_{Mor} \right) \rightarrow y$.

\subsection{Hierarchical Interaction Framework }\label{Sec.1}

\noindent\textbf{Multimodal Interaction Module. }Although fine-grained subgraphs better characterize the regional and modular properties of brain networks, they are more prone to noise than global connectome and computational costs. To effectively capture interactions within subgraphs, we proposed a Siamese GNN framework with a cross-modal attention mechanism. Given the $j^{th}$ subgraphs $G_{S}^{j}$ and $G_{Mor}^{j}$ extracted from SC and MC, respectively, a Siamese GNN first learns embeddings for both:

\begin{equation}
    Z_{S}^{j}\ =\ \varPhi \ \left( G_{S}^{j},\ \theta \right),  
    Z_{Mor}^{j}\ =\ \varPhi \ \left( G_{Mor}^{j},\ \theta \right), 
\end{equation}
where $Z_{S}^{j}\in \mathbb{R}^{|V_j|\times d}$ and $Z_{Mor}^{j}\in \mathbb{R}^{|V_j|\times d}$ are node embeddings for both types of subgraphs, and $\theta $ represents the shared parameters of the GNN encoder.

A cross-modal attention mechanism further integrates information from both modalities. Based on the biological prior that SC provides a stable framework reflecting FC, we define the query (Q), key (K) and value (V) matrix as follows:

\begin{equation}
    Q_{j}\ =\ W_{Q}Z_{Mor}^{j},\ K_{j}\ =\ W_{K}Z_{S}^{j},\ V_{j}\ =\ W_{V}Z_{S}^{j},
\end{equation}
where $W_Q,W_K,W_V\in \mathbb{R}^{d\times d}$ are learnable weight matrices. Then the fusion embedding of the $j^{th}$ subgraph $Z_j$ is computed as follows:
\begin{equation}
    Z_j \ =\ soft\max \left( \frac{Q_{j}\left( K_{j} \right) ^T}{\sqrt{d_k}} \right) V_{j}^{h},
\end{equation}
where $d_k$ is the dimensionality of the key vector.

\noindent\textbf{Regional Interaction Modeling. }The receptive field, comprising a brain region (node) $i$  and its neighbors, reflects regional connectome characteristics. Thus, the regional interaction representation $z_{r}^{i}$ of node $i$ is calculated as:

\begin{equation}
Z_{r}^{i}\ =\ \mathrm{MIM}\left( G_{S}^{i},\ G_{Mor}^{i} \right), 
\end{equation}

where $G_{S}^{i}$ is the regional subgraph consisting of node $i$ and its multi-hop neighbors within SC, while $G_{Mor}^{i}$ is the regional subgraph derived from corresponding MC.  Inspired by~\cite{ye2023rh}, each regional subgraph $G_r^{i}$ is comprised of node $i$ and its one-hop neighbor captured through a linear layer and softmax.

\noindent\textbf{Modular Interaction Modelling. }A modular subgraph $G_{m}$ consists of multiple nodes with similar semantics, reflecting the modular characteristics of the brain networks. We capture modular subgraphs of each type of brain connectome by feeding regional interaction representations $Z_{r}$ into the PMP module. Then the modular interaction representation can be obtained as follows:

\begin{equation}
Z_{m}^{k}\ =\ \mathrm{MIM}\left( G_{s}^{k},\ G_{Mor}^{k} \right), 
\end{equation}
where $Z_{m}^{k}$ is the modular interaction representations of the $k^{th}$ module. $G_{S}^{k}$ and $G_{Mor}^{k}$ represent the $k^{th}$ modular subgraph from SC and MC, respectively.

Finally, the global interaction representation $Z_{g}$ can be obtained by fusing representations from regional and modular scales through the MFF mechanism. We then feed it into an MLP to capture $Z_g$ for downstream predictions.


\subsection{Multiscale Features Fusion }\label{Sec. 2}
Compared to modular representations, regional representations contain higher redundancy. Simple concatenation of these features may introduce redundant information, thereby constraining the expressive power of global representations. To address this limitation, we implement a Multiscale Features Fusion mechanism that enables effective fusion of regional and modular features.

Based on the prior that modular representations are more compact and refined than regional representations, we leverage modular representations to filter out redundant features from regional representations adapted from Zhao's method\cite{zhao2019deep} for features purification in signals. Firstly, the modular representation $Z_m$ is mapped to regional scale via two linear layers, yielding clean regional representation $Z_{m'}$. Global Average Pooling layer then compress and aggregates features within each region into a vector $T$. Subsequently, $T$ is processed by an MLP followed by a sigmoid activation function to produce a scaling coefficient $a$, which adaptively modulates the threshold. The final soft threshold $\tau$ is computed by multiplying $a$ and $T$. This threshold $\tau$ is applied to the original regional representation $Z_r$, eliminating redundant features below $\tau$. Finally, the purified regional representation $Z_{r'}$ and original modular representation $Z_m$ are concatenated and averaging to generate the global representation $Z_g$:
\begin{equation}
\begin{aligned}
& Z_{m'} = \text{Linear}(Z_m),\quad 
T = \text{GAP}(Z_{m'}),\quad 
a = \text{Sigmoid}(\text{MLP}(T)),\quad 
\tau = a T \\
& \hspace{8em} Z_{r'} = \begin{cases} 
Z_r - \tau, & Z_r > \tau \\
0, & -\tau \leq Z_r \leq \tau \\
Z_r + \tau, & Z_r < -\tau 
\end{cases} \\[2ex]
& \hspace{8em} Z_g = \text{mean}\Big(\text{Concatenate}\big(Z_{r'}, Z_m\big)\Big)
\end{aligned}
\end{equation}

\subsection{Personalized Modular Partition }\label{Sec. 3}
We propose a soft module assignment based on regional interaction representation, inspired by~\cite{ying2018hierarchical}. Since the modularity of brain connectome reflects cross-region morphological connectivity\cite{wang2024toward}, we obtain the soft module assignment matrix $S$ with regional interaction embeddings $Z_r$ and $G_{Mor}$:


\begin{equation}
    S=\mathrm{Thresh}\left( \mathrm{softmax} \left( \mathrm{GNN}\left( Z_r,G_{Mor} \right) \right) \right).
\end{equation}
A threshold $t$ is applied to the $S$ matrix to achieve modular partition. To further improve the relevance of modular partition, we improve the modularity score $q$ ~\cite{sporns2016modular} originally calculated solely based on MC topology by replacing it with a combined framework of SC and MC. Based on this improved modularity score $q$, we design our loss:





\begin{equation}
L_q=-q_s=-\sum_{u=1}^2{\frac{1}{2w}Tr\left( S^T\left( A-\frac{bb^T}{2w} \right) S \right)},
\end{equation}
where $A$ is the modularized adjacency matrix, and $u$ is the type of connectome. $b$ is the node degree vector calculated from $A$, and $w=\frac{1}{2}\sum{A_{ij}}$ is the total edge weight.
Finally, the summarized loss function ($L_{total}$) can be expressed as:
\begin{equation}
L_{total}=\eta _1L_{task}+\eta _2L_q,
\end{equation}
where $L_{task}$ is the cross-entropy loss, which is utilized for classification tasks. The $\eta_1$ and $\eta_2$ are loss weights.

\section{Experiments and Results}
\subsection{Datasets and Data Preprocessing}
We used the UCSF-PDGM dataset\cite{calabrese2022university} for model evaluation. It includes patients with gliomas ranging from Grade 1 to Grade 4, comprising 103 IDH-mutant cases and 392 IDH-wildtype cases. The UCSF-PDGM dataset comprises multi-modal MRI including T1, T2, and fractional anisotropy (FA) sequences. For our analysis, we specifically utilize FA and T1 contrast-enhanced (T1c) MRI modalities to construct SC and MC. Patients were divided into training, validation and testing sets with a ratio of \text{7:1:2}.

To achieve more robust characterization of structural connectivity networks in glioma patients, we adopted the methodology established by~\cite{wei2021quantifying}. Specifically, we estimated the connectivity strength between each pair of brain regions in glioma patients by combining an unbiased tractography template constructed from healthy dMRI data using the IIT template\cite{varentsova2014development} and Destrieux atlas\cite{destrieux2010automatic} with patient-specific FA skeletons generated by TBSS in FSL. We utilized the R2SN library\cite{zhao2022regional} to generate morphological connectivity networks by first registering T1c images to MNI152 space, extracting regional morphological features using PyRadiomics, and computing inter-regional similarities via Pearson correlation.


\subsection{Experiment Setup and Evaluation}
Our experiments were conducted using PyTorch and Pytorch Geometric on an NVIDIA L20 GPU with 48GB of memory. The performance was compared by accuracy and F1 score for IDH prediction task. The learning rate is set as 1$\times 10^{-4}$ for the Adam Optimizer. The
loss weights (i.e., $\eta_1$, $\eta_2$) were set as 0.5 and 0.2 respectively. For comprehensive comparisons, we include: four baselines of single-modal graph learning (i.e., MLP, GraphSage~\cite{hamilton2017inductive} and DiffPool~\cite{ying2018hierarchical}), and their multimodal version. We also compare three SOTA methods combining structural and functional connectomes(i.e., Cross-GNN\cite{yang2023mapping}, sf-GBT\cite{peng2025joint} and Joint-GCN~\cite{li2022joint}), representing the integration methods.

\begin{table}[t]
\centering
\setlength{\tabcolsep}{15pt} 
\fontsize{8}{10}\selectfont
\renewcommand{\arraystretch}{0.9}
\caption{IDH prediction on UCSF-PDGM dataset. \textbf{bold} indicates the best results.}
\label{Tab:1} 
\begin{tabular}{cccc}
\toprule
\multirow{2}{*}{\textbf{Modality}} & \multirow{2}{*}{\textbf{Method}} & \multicolumn{2}{c}{\textbf{IDH}} \\ 
\cmidrule{3-4} 
 &  & \textbf{ACC~$\uparrow$} & \textbf{F1~$\uparrow$} \\
\midrule
\multirow{3}{*}{SCN} 
 & MLP & 70.71 & 12.12 \\
 & GraphSage & 73.74 & 31.58 \\
 & DiffPool & 72.73 & 22.86 \\
\midrule
\multirow{3}{*}{R2SN} 
 & MLP & 71.72 & 26.32 \\
 & GraphSage & 68.69 & 20.51 \\
 & DiffPool & 66.67 & 35.29 \\
\midrule
\multirow{7}{*}{Multi-modal} 
 & MLP & 71.72 & 36.36 \\
 & GraphSage & 74.75 & 52.83 \\
 & DiffPool & 72.73 & 34.15 \\
 & Cross-GNN & 73.74 & 59.38 \\
 & Joint-GCN & 78.79 & 63.16 \\
 & sfBGT & 81.82 & 66.67 \\
 & Ours & \textbf{84.85} & \textbf{72.73} \\
\bottomrule
\end{tabular}
\end{table}

\begin{table}[t]
\centering
\setlength{\tabcolsep}{11pt} 
\fontsize{9}{11}\selectfont
\renewcommand{\arraystretch}{1.1}
\caption{Effectiveness of each module on UCSF-PDGM dataset for IDH prediction.}
\label{Tab:2} 
\begin{tabular}{cccccc}
\toprule
\multirow{2}{*}{\textbf{MIM}} & \multirow{2}{*}{\textbf{PMP}} & \multirow{2}{*}{\textbf{MFF}} & \multicolumn{2}{c}{\textbf{IDH}} \\ 
\cmidrule{4-5} 
 & & & \textbf{ACC~$\uparrow$} & \textbf{F1~$\uparrow$} \\
\midrule
$\checkmark$ & & & 71.72 & 50.00 \\ 
 & $\checkmark$ & & 78.79 & 66.67 \\ 
 &  & $\checkmark$ & 84.58 & 69.39 \\
$\checkmark$ & $\checkmark$ & $\checkmark$ & \textbf{84.85} & \textbf{72.73} \\
\bottomrule
\end{tabular}
\end{table}

\subsection{Model Comparisons}
 As shown in Table \ref{Tab:1}, on the UCSF-PDGM dataset, our proposed framework achieves the best performance. Specifically, our model  surpasses the best multimodal method by 3.03\% in accuracy and 6.06\% in F1 score. In general, multimodal methods outperform unimodal methods, which conclusively demonstrates the effectiveness of structure-radiomics integration. The superiority of our framework could be attributed to our advantage of properly capturing the interactions between multimodal brain connectomes.

\subsection{Ablation Studies}
 To assess the effectiveness of each module, we compare the baseline with three variants: 1) ablating the Multimodal Interaction Module; 2) ablating the Personalized Modular Partition module; 3) ablating the Multiscale Features Fusion mechanism. As shown in Table \ref{Tab:2}, the full model outperforms all the variants on the UCSF-PDGM dataset, indicating the effectiveness of each module.


\section{Conclusion}
This study introduces a hierarchical framework integrating structural connectome and morphological connectome for IDH mutation status prediction. Our framework includes a multimodal interaction module based on a Siamese network and cross-modal attention, a multiscale features fusion mechanism inspired by soft thresholding in signal processing, and a personalized modular partitioning method based on the modularity score. Our framework outperforms competing methods in predicting IDH status in UCSF-PDGM datasets, demonstrating superior performance in multimodal brain connectome analysis. Future studies will further develop the model to identify salient hierarchical interactions of structural connectome and morphological connectome in the brain of tumor patients.

\bibliographystyle{splncs04}
\bibliography{MICCAI}
\end{document}